\documentclass[10pt, a4paper, twocolumn]{article}
\usepackage[a4paper,
           rmargin=1in, 
           top=1cm]{geometry}
\usepackage{graphicx}
\usepackage{mathrsfs}
\usepackage{amsfonts}
\usepackage{amsmath}
\usepackage{lipsum} 
\usepackage[table]{xcolor}
\usepackage{booktabs}
\usepackage{tabularx}

\usepackage{titlesec}

\titleformat{\section}
  {\bfseries\fontsize{12pt}{15pt}\selectfont} 
  {\thesection}
  {0.4em}                          
  {}

\titleformat{\subsection}
  {\bfseries\fontsize{10pt}
  {5pt}\selectfont} 
  {\thesubsection}
  {0.4em}                          
  {}

\begin{document}  

\title{Knowledge Trees: Gradient Boosting Decision Trees on Knowledge Neurons as Probing Classifier}

\author{S.A. Saltykov}

\maketitle


\noindent\hfill\begin{minipage}{0.85\linewidth}
\textbf{\fontsize{12pt}{15pt}\selectfont Abstract}\vspace{0.5em}

To understand how well a large language model captures certain semantic or syntactic features, researchers typically apply probing classifiers. However, the accuracy of these classifiers is critical for the correct interpretation of the results. If a probing classifier exhibits low accuracy, this may be due either to the fact that the language model does not capture the property under investigation, or to shortcomings in the classifier itself, which is unable to adequately capture the characteristics encoded in the internal representations of the model. Consequently, for more effective diagnosis, it is necessary to use the most accurate classifiers possible for a particular type of task. Logistic regression on the output representation of the transformer neural network layer is most often used to probing the syntactic properties of the language model.

We show that using gradient boosting decision trees at the Knowledge Neuron layer, i.e., at the hidden layer of the feed-forward network of the transformer as a probing classifier for recognizing parts of a sentence is more advantageous than using logistic regression on the output representations of the transformer layer. This approach is also preferable to many other methods. The gain in error rate, depending on the preset, ranges from 9-54\%.

\end{minipage}\hfill\null

\section{Introduction}

Hallucinations are currently a serious problem of large language models\cite{TransformersAndCompositionality}. It is becoming clear that straightforward attempts to solve this problem (such as increasing model size, training data, and cleanliness) are not only costly, but also not effective enough: large language models continue to hallucinate at an unacceptably high frequency \cite{Hallucination}.

\begin{figure}[ht]
  \centering
  \begin{minipage}{\columnwidth}
    \includegraphics[width=\linewidth]{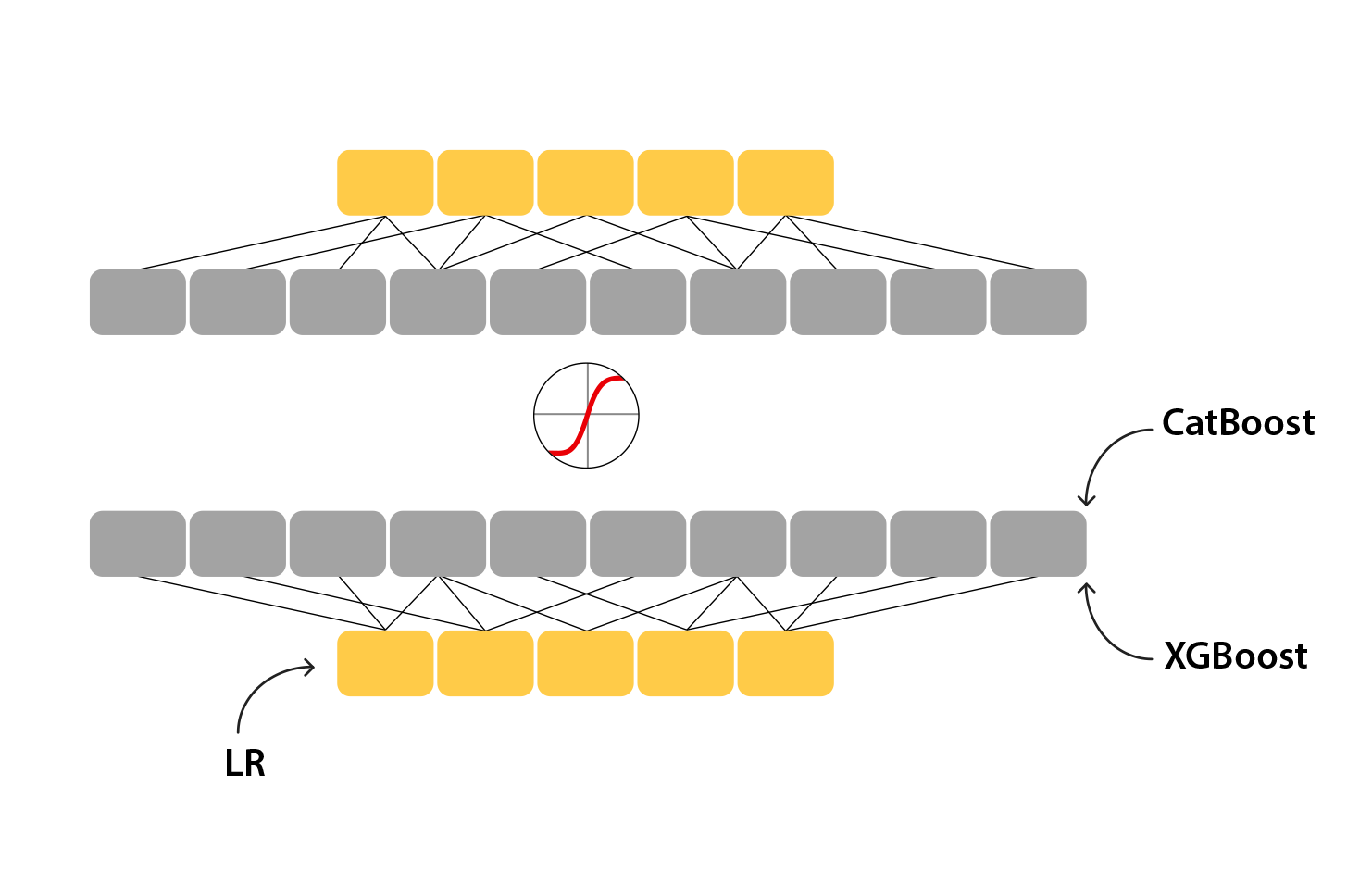}
    \caption{Knowledge Trees (schematic diagram). Gradient Boosting Decision Trees probes Knowledge Neurons. In contrast to the more traditional approach, where logistic regression (LR) probes the vector representation of a token.}
  \end{minipage}
\end{figure}

One potential solution to this problem could be the creation of a hybrid model, in which a Large Language Model (LLM) is combined with a knowledge graph \cite{LLMandKG}. To obtain the knowledge graph, it can either be directly extracted from the language model or it can be obtained through a multi-step process---extracting entities and relations and then selecting the graph that best matches the identified entities and relations.

In the case of syntactic parsing, the extraction of entities is found to be closely related to the use of probing classifiers \cite{Belinkov}.

Thus, it emerges that one of the crucial steps, particularly in addressing hallucinations, is the selection of an appropriate type of probing classifier that allows for the optimal identification of syntactic entities.

This paper explores the use of gradient boosting decision trees on the hidden layers of transformer neural networks for probing classifiers. The study aims to improve the process of understanding and interpreting the capabilities of large language models (LLMs) in capturing syntactic features. Traditional probing methods like logistic regression often face accuracy limitations, making it difficult to determine whether a language model truly captures a specific property or whether the shortcomings lie in the classifier. 

This research proposes using gradient boosting decision trees at the Knowledge Neuron layer, demonstrating that this method is more effective than logistic regression, with error rate improvements ranging from 9\% to 54\%. The paper includes detailed experimental analysis, exploring various scenarios and comparing the effectiveness of different probing classifiers, including logistic regression, multilayer perceptron, support vector machines, naive bayes, and others. 

The results indicate that Knowledge Trees, especially when applied to complex syntactic structures on medium-sized datasets, show significant advantages over traditional methods.

\section{Related works}

In analyzing the extent to which a large language model can capture a certain syntactic property, the de-facto standard involves training a logistic regression classifier on vector representations from approximately the middle layers \cite{Belinkov, 7layer}. Typically, the type of regularization used in training the classifier is not specified; the default option for logistic regression is either to use l2-regularization or not to use any regularization at all. Therefore, we will choose this as our baseline – a logistic regression with l2-regularization, trained on vector representations from roughly the middle layer of the transformer.

\subsection{Probing classifiers and syntactic entities}

Probing classifiers are designed to determine whether some semantic and/or syntactic quality is captured by a trained model (most often a neural network) and, if so, where in that model it is captured. In the case of transformer neural networks \cite{Attention} for NLP tasks, we are talking about representations of a token that change as the token passes through the transformer. In this paper, we focus on syntactic properties, specifically probing parts of a sentence. 

In order to make this task the least language-specific (as much as possible), we have chosen the Universal Dependencies dataset \cite{UD}. It introduces a nomenclature of 37 parts of a sentence that are common to more than 100 languages. In the Universal Dependencies dataset, we selected ``en\_lines'' sub-dataset. Further we analyzed only those parts of a sentence that are found in both training and test sample of this dataset. There are 32 such parts of a sentence, and they occur in sentences with significantly different frequencies. This information is summarized in Table 1.

The ``en\_lines'' dataset has 3176 sentences in the training sample and 1035 sentences in the test sample. In train sample, for example, punctuation marks occur 6679 times, while the part of a sentence ``iobj'' occurs only 49 times. This determines the number of positive examples when training the classifier to probe this part of a sentence. We will consider all other tokens in the same sentence as negative examples. Obviously, there are many times more negative examples (and most often several orders of magnitude more) than positive examples. 

Different machine learning methods inherently perform unequally well on unbalanced datasets. To keep the comparison clean, we make the dataset for each part of a sentence balanced by randomly duplicating positive examples until their number equals the number of negative examples. The number of rows in the resulting training balanced datasets is also shown in Table 1. In addition, the table displays the accuracy on the test balanced dataset achieved by the logistic regression (trained on the training balanced dataset).

\begin{table}[ht]
\begin{tabularx}{\columnwidth}{l>{\hsize=0.5\hsize}Xrp{1.5cm}p{1.2cm}} 
\toprule
{} &  taxon &  positive &  rows &        LR \\
\midrule
1  &      orphan &         2 &              66 &  0.500 \\
2  &        iobj &        49 &            2526 &  0.927 \\
3  &    vocative &        69 &            1652 &  0.983 \\
4  &  dislocated &        86 &            4310 &  0.661 \\
5  &       csubj &       112 &            4764 &  0.831 \\
6  &   discourse &       118 &            3604 &  0.943 \\
7  &        expl &       197 &            8694 &  0.998 \\
8  &   parataxis &       252 &            9908 &  0.804 \\
9  &      nummod &       280 &           12854 &  0.894 \\
10 &       fixed &       292 &           14036 &  0.895 \\
11 &       appos &       340 &           12834 &  0.806 \\
12 &       ccomp &       359 &           16202 &  0.776 \\
13 &        flat &       435 &           13152 &  0.927 \\
14 &       xcomp &       888 &           33936 &  0.871 \\
15 &         cop &       929 &           33888 &  0.976 \\
16 &       advcl &      1006 &           42492 &  0.893 \\
17 &         acl &      1125 &           46018 &  0.880 \\
18 &          cc &      1715 &           67468 &  0.993 \\
19 &    compound &      1788 &           45460 &  0.895 \\
20 &        mark &      1795 &           61062 &  0.977 \\
21 &         aux &      1950 &           62292 &  0.992 \\
22 &        conj &      2747 &           69320 &  0.917 \\
23 &         obj &      2779 &           76814 &  0.931 \\
24 &      advmod &      3057 &           77250 &  0.958 \\
25 &        amod &      3158 &           71602 &  0.959 \\
26 &        root &      3270 &          122458 &  0.944 \\
27 &         obl &      3373 &           84772 &  0.913 \\
28 &        nmod &      3805 &           83168 &  0.906 \\
29 &         det &      5131 &           97580 &  0.997 \\
30 &       nsubj &      5303 &          111322 &  0.962 \\
31 &        case &      5398 &           98704 &  0.987 \\
32 &       punct &      6679 &          113738 &  0.9995 \\
\bottomrule
\end{tabularx}
\caption{Datasets for training probing classifiers for different sentence members. The short name of the sentence member according to \cite{UD}, the number of unique positive examples, the total number of examples in the balanced dataset, and the accuracy of the logistic regression trained on this dataset are given.}
\end{table}

\subsection{Transformer architecture and probing objects}

During sentence processing, each word (or, more precisely, token is a word, part of a word or punctuation mark) is passed through a transformer neural network. A transformer \cite{Attention} consists of a stack of transformer blocks. Each such block, or in another way a transformer layer, consists of two components: a self-attention block and a fully connected neural network with one hidden layer. 

The self-attention block is designed to enrich the token representation vector with the representations of neighboring tokens. Depending on the representation properties of the current token, different neighboring tokens are given different attention during its enrichment. That is, the self-attention block receives the representation vector of the current token as input and outputs the representation vector of the current token enriched with information from neighboring tokens.

The block of a fully-connected neural network with one hidden layer, in its turn, also receives as input a vector of representation of the current token and outputs a vector of representation of the current token enriched with insights that logically follows \cite{KeyValueMemory, SubUpdates} from this representation and does not take into account representations of neighboring tokens in any way. The token representation is then passed to the next transformer layer, which has the same structure but different weights. 

In order to understand whether the current representation reflects that this token is such a part of the sentence, one can probe any of the token representations obtained along the way in the transformer. The initial embeddings of tokens, the output representations of any of the transformer layers, and any intermediate representations within the transformer layer can be probed. Most often the output representations of the transformer layers are probed. But which depth layer to choose? 

Quite often the last, output layer of the transformer is chosen for probing. This is because the representation vector is most enriched on the last layer, and if any layer captures that this token is such and such a part of a sentence, it seems that this layer is the last, output layer. However, there may be a problem here: the more enriched with various aspects from neighboring tokens the current token becomes, the more likely it is to be overfitted, especially on small datasets. And since syntactic markup of text is manual and complex, syntactic markup datasets are likely to be relatively small. To avoid overfitting probing classifiers, it is often advisable to choose a transform layer at about the middle or two-thirds of the depth. Experiments show that this depth turns out to be sufficient to capture the syntactic essence, however, the token representation is not yet enriched enough to lead to significant overfitting.

Thus, a kind of ``gold standard'' of probing has emerged, where the syntactic entity is detected on the output vector of the transformer layer representation at about two-thirds of the transformer depth.

However, as we noted above, it is possible to probe not only the output representation vector of some transformer layer, but also the intermediate representations that occur within the transformer layer.  Intermediate representations of a token arising inside a self-attention block as a probing object is an interesting and promising subject of a separate study; we will not touch upon it in this paper. But we will consider intermediate representations inside the feed-forward network block (FFN-block) in more detail.

Since the FFN-block in the transformer is a fully connected neural network with one hidden layer, the intermediate representation of the token is a vector of this hidden layer. It can also be the object of probing. Interestingly, studies show that this FFN hidden layer vector as an alternative to output representation of the transformer layer as a way to encode the token, differs significantly from the output representation in terms of interpretive capabilities. It has been shown that FFN's hidden layer can be viewed as knowledge neurons \cite{KnowledgeNeurons}. That is, many components of this FFN'a hidden layer vector individually encode some real world object or phenomenon. In contrast, this is not the case for the original token embeddings and for the output representations of the transform layer. 

It is worth noting that the initial embeddings and output representations are, as it were, in the same vector space, and the enrichment of the source token embedding as one moves through the transformer is a shift of a point in space as the context is more fully considered. Nevertheless, the individual coordinates of this vector space are extremely weakly interpretable. The entire set of coordinates (i.e., a point) or a subspace in some neighborhood of a point is quite often quite well interpreted, while the individual coordinates seem to have no ``physical'' meaning.

On the contrary, the individual coordinates of the point into the vector space of the FFN's hidden layer, in other words, the individual components of the FFN's hidden layer vector are quite well interpreted. Thus, this difference from a meaningful point of view between the intermediate representation of the FFN and the output representation of the transformer layer makes the former also an interesting probing object to study.

Furthermore, we can go further and realize that we can use not only those intermediate representations that actually occur when a vector passes through the transformer, but also form our own representation, by analogy with the intermediate ones that actually occur. We can consider that from the vector space of embeddings (in which a point moves when enriching the vector representation of a token) is translated into the vector space of knowledge neurons by multiplication by an appropriate matrix of weights, which for simplicity and uniformity in the following we will call FFN-matrix. Then the bias vector is added and after that the activation function is applied. In principle, nothing prevents you from experimenting with these transformations: to include or not to include them in the set of transformations, as well as to multiply the output representations of some layers by FFN-matrices of completely different layers. And from all combinations of these possibilities, choose the one that achieves the best result. 

Our preliminary calculations (not included in this paper) show that for logistic regression, taking information from the second-to-last layer is the most preferable. This is quite consistent with existing results about where syntactic properties are most expressed (when probed by logistic regression). Therefore, we will choose the fifth, second-to-last layer as the output layer representations. 

Voluntaristically, we will choose the FFN matrix from the fourth layer. Preliminary calculations show that this works well, but a more reasonable choice of the transform layer for the FFN matrix is the subject of a separate study. Also, adding the bias vector does not provide an advantage for gradient boosting, so we will omit this operation. Also preliminary calculations show that application of activation function significantly reduces the performance, so we will not apply it after multiplication by FFN-matrix before probing.

So, let us choose two probing objects for our study. The first is the output representation of the second-to-last, the fifth layer. The second is the same output representation multiplied by the FFN-matrix from the fourth layer.

\section{Experiment design}

\subsection{Knowledge Neurons being probed}

Multiplication of a vector by a matrix can be interpreted as a translation of a point from one vector space to a point in another vector space. This is a rather useful interpretation, including for this study. But it is possible to interpret this multiplication in another way. A matrix is a set of vectors. Therefore, multiplication of a vector by a matrix is a set of multiplications of a vector by a vector. If vectors are normalized, multiplication of a vector by a vector can be interpreted as determining their degree of similarity. But it can also be interpreted in another way. A vector uniquely defines a hyperplane perpendicular to it and passing through its end. Multiplication of a vector by a vector defines the projection of one vector onto the hyperplane defined by the second vector. 

For example, one Knowledge Neuron is the result of taking the product of the token representation vector and one of the vectors in the FFN matrix. In other words, the value of a Knowledge Neuron is the projection of the representation vector onto the hyperplane defined by the vector from the FFN matrix. If a particular Knowledge Neuron correlates reasonably well with some class of phenomena, then mathematically this will mean that the ends of the vectors of tokens denoting (in context!) the given phenomenon are on one side of the hyperplane corresponding to the vector from the FFN-matrix, and all other phenomena are on the other side of the hyperplane.

Multiplication of the vector by the FFN-matrix gives a correspondence of the end of the vector with a whole set of hyperplanes: on which side of them the end of the vector is located and at what distance. Such a set of projections onto well-chosen hyperplanes can very accurately characterize the projected vector. A verbal analogy is suggested here: in classifying this vector, we rely on a set of already pre-trained, ``supporting'' hyperplanes that convey some syntactic and/or semantic characteristics well. 

Therefore, in principle, the vector of representations can be multiplied by the FFN-matrix from any layer, the ``physical meaning'' will be the same. It is only important to choose the FFN-matrix, i.e., the set of hyperplanes that most accurately reflects the property under study. In other words, the set of hyperplanes on which it is most reasonable to ``rely'' when trying to comprehend a given syntactic quality.

\subsection{Analysed machine learning methods}

We choose the following machine learning methods to analyze in order to understand how good probing classifiers can be constructed on their basis. First of all, we will take methods that are traditionally ``friendly'' to neural network architectures: logistic regression and multilayer perceptron (MLP). For LR and MLP we will take the default Sklearn implementation. Then we take tree-based machine learning methods: Random forest and Gradient Boosting. Our hypothesis is that tree-based methods have a chance to overfit less on small datasets than methods trained by back propagation. For completeness, we also take methods from ``classical'' machine learning: $k$-Nearest neighborhood, Naive Bayes classifier and Support Vector Machine. Regarding the latter, we have no clear hypotheses as to can they  outperform logistic regression or not in this class of problems, but nevertheless it is worth testing them as well.

Our main hypothesis is that tree-based machine learning methods in the form of gradient boosting have the highest chances to compete with logistic regression, so we will take not only the default Sklearn implementation, but several variants of gradient boosting. We will consider implementations by the Sklearn, XGBoost \cite{XGBoost} and CatBoost \cite{CatBoost} libraries, which differ not only in program code but also in the regularizations used. In addition, in the ensemble of trees built by gradient boosting we will consider all tree depths from 1 to the default depth (which is 3 for Sklearn and 6 for XGBoost and CatBoost). Thus, we will explore 21 variants of machine learning methods. We apply each of these methods to two probing objects and thus obtain 42 explored probing classifiers. This information is summarized in the Table 2.

\subsection{Knowledge Trees}

So, we have 15 variants of gradient boosting (6+6+3) and each of them can be applied to Knowledge Neurons. \textit{The gradient boosting applied to Knowledge Neurons will be called Knowledge Trees}. Our main hypothesis is that gradient boosting by itself helps reduce overfitting, and that applying it to Knowledge Neurons specifically is more native to it, and therefore effective, than to the output representation of the transformer layer. Thus, of the 42 probing classifiers, we categorize 15 of them as Knowledge Trees.

We chose DistilBERT \cite{DistilBERT} as the LLM through which text is passed and within which vector representations of tokens and Knowledge Neurons, which are the objects of probing, are respectively obtained.

We did not analyze logistic regression with L1-regularization (although this may be a promising topic for a separate study), since its existing implementations are not very robust in terms of speed of execution and memory consumption \cite{l1Unpredictability}.

In this experiment, we employ 42 distinct probing classifiers to predict 32 different types of parts of a sentence. Each of these parts of a sentence is predicted using all 42 probing classifiers. Essentially, every probing classifier addresses the same 32 classification tasks, allowing for a comparative analysis of method effectiveness on each task. Performance varies; some probing classifiers excel in certain tasks, while others perform better in different ones. Despite occasional underperformance, a classifier can sometimes be statistically more significant than its counterparts.

\begin{table}[ht]
\begin{tabularx}{\columnwidth}{l>{\hsize=0.4\hsize}Xp{0.8cm}p{1cm}} 
\toprule
{} &                 methods & repr. & neurons \\
\midrule
1  &     Logistic regression &            LR &      kn\_LR \\
2  &   Multilayer perceptron &           MLP &     kn\_MLP \\
3  &           Random forest &            RF &      kn\_RF \\
4  &    k-Nearest neiborhood &           KNN &     kn\_KNN \\
5  &  Support vector machine &           SVC &     kn\_SVC \\
6  &             Naive bayes &            NB &      kn\_NB \\
7  &      Sklearn GB, 3 lev. &          GB\_3 &    kn\_GB\_3 \\
8  &      Sklearn GB, 2 lev. &          GB\_2 &    kn\_GB\_2 \\
9  &      Sklearn GB, 1 lev. &          GB\_1 &    kn\_GB\_1 \\
10 &          XGBoost, 6 lev &         XGB\_6 &   kn\_XGB\_6 \\
11 &          XGBoost, 5 lev &         XGB\_5 &   kn\_XGB\_5 \\
12 &          XGBoost, 4 lev &         XGB\_4 &   kn\_XGB\_4 \\
13 &          XGBoost, 3 lev &         XGB\_3 &   kn\_XGB\_3 \\
14 &          XGBoost, 2 lev &         XGB\_2 &   kn\_XGB\_2 \\
15 &          XGBoost, 1 lev &         XGB\_1 &   kn\_XGB\_1 \\
16 &         CatBoost, 6 lev &          Ct\_6 &    kn\_Ct\_6 \\
17 &         CatBoost, 5 lev &          Ct\_5 &    kn\_Ct\_5 \\
18 &         CatBoost, 4 lev &          Ct\_4 &    kn\_Ct\_4 \\
19 &         CatBoost, 3 lev &          Ct\_3 &    kn\_Ct\_3 \\
20 &         CatBoost, 2 lev &          Ct\_2 &    kn\_Ct\_2 \\
21 &         CatBoost, 1 lev &          Ct\_1 &    kn\_Ct\_1 \\
\bottomrule
\end{tabularx}
\caption{Investigated probing classifiers by what machine learning methods they use and where exactly the probing is done.}
\end{table}

This leads to three potential scenarios: the first probing classifier is statistically significantly more accurate than the second, the second probing classifier is statistically significantly more accurate than the first, or neither probing classifier demonstrates a statistically significant advantage over the other.

Based on this, we can construct a dominance graph on the set of probing classifiers. By removing edges that derive from transitivity, we achieve a more refined dominance graph. This graph will be illustrated in the figures provided.

\section{Results}

In this section, we will show that two factors are crucial for the composition of the non-dominated set of probing classifiers: the complexity of the classification task, as measured by the accuracy of the logistic regression on the dataset corresponding to the part of the sentence, and the size of the training dataset. Therefore, the graph construction experiments described earlier will be conducted in five different setups.

\subsection{General case}

When visualizing the graph, we make sure that all elements of the non-dominated set are present at the top level, and only them. We don't pay much attention to the mutual location of the other elements of the graph relative to each other, they are automatically drawn by the library used for graph visualization.

\begin{figure}[ht]
  \centering
  \begin{minipage}{\columnwidth}
    \includegraphics[width=\linewidth]{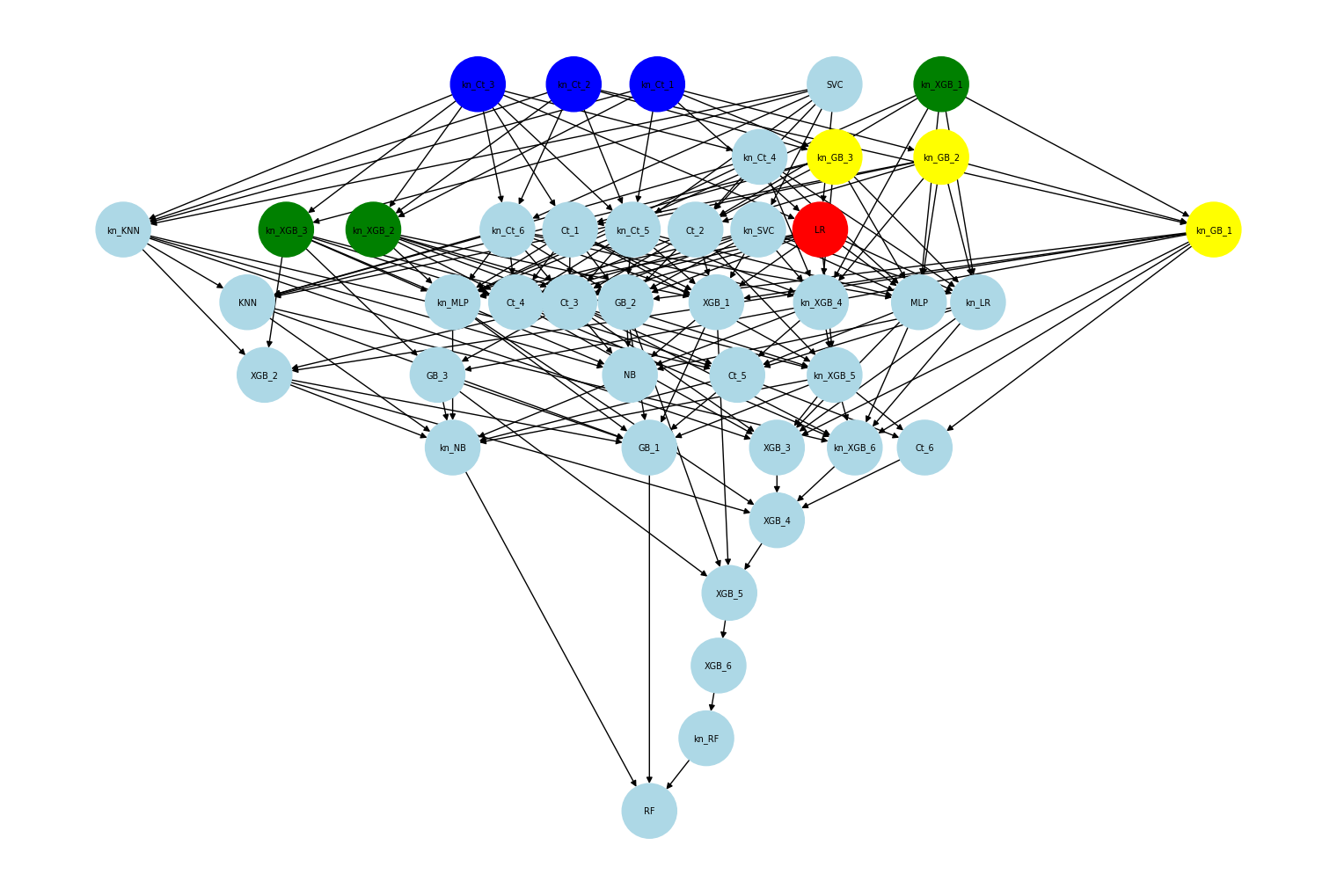}
    \caption{Dominance graph of probing classifiers without dataset constraints for different sentence members. Enlarge the .pdf file to read the names of the classifiers. CatBoost-based classifiers are blue, XGBoost-based classifiers are green, and Sklearn-based classifiers are yellow. The probing classifier based on logistic regression is red.}
  \end{minipage}
\end{figure}

In the general case, i.e., without imposing additional constraints, Figure 2 allows us to make the following conclusions. The non-dominated set includes methods from the Knowledge Trees group, as well as the Support Vector Machine (SVC) classifier. Logistic Regression (LR), as we see, is the dominated method.

It is also important to note that the non-dominated set of probing classifiers from the Knowledge Trees group includes three CatBoost-based methods, one XGBoost-based method, and none Sklearn-based. The absence among the leaders of ``standard'' gradient boosting without specific regularizations seems reasonable. However, the fact that more CatBoost-based methods than those based on the better-known XGBoost are among the non-dominated ones is an intriguing point for further analysis to understand why this is the case.

\subsection{Relatively difficult classification tasks}

We calculate the rank correlation between the accuracy of logistic regression and the accuracy advantage that methods from the Knowledge Trees group have over logistic regression in probing 32 distinct part of sentence (Table 3).

\begin{table}[ht]
\centering
\begin{tabularx}{\columnwidth}{p{3.3cm}rr}
\toprule
Method & Correlation & p-value \\ 
\midrule
\rowcolor{gray!30} \textbf{kn\_Ct\_1} & \textbf{-0.58} & \textbf{0.0005} \\ 
kn\_Ct\_2 & -0.25 & 0.1679 \\ 
kn\_Ct\_3 & -0.02 & 0.9001 \\
\rowcolor{gray!30} \textbf{kn\_XGB\_1} & \textbf{-0.44} & \textbf{0.0124} \\ 
kn\_XGB\_2 & -0.13 & 0.4728 \\ 
\rowcolor{gray!30} \textbf{kn\_XGB\_3} & \textbf{0.50} & \textbf{0.0033} \\ 
\rowcolor{gray!30} \textbf{kn\_GB\_1} & \textbf{-0.56} & \textbf{0.0008} \\ 
\rowcolor{gray!30} \textbf{kn\_GB\_2} & \textbf{-0.54} & \textbf{0.0013} \\ 
kn\_GB\_3 & -0.33 & 0.0618 \\ 
\bottomrule
\end{tabularx}
\caption{Rank correlation between the advantage in accuracy of Knowledge Trees probing classifiers over Logistic Regression (LR) accuracy and LR accuracy in probing different sentence members.}
\label{your-label}
\end{table}

We observe at least two patterns. First, for the shallow depth methods in the Knowledge Trees group, the less accurately the logistic regression probes a part of the sentence, the greater the advantage Knowledge Trees give. This result is statistically significant for all three methods at depth 1 and for one method (kn\_GB\_2) at depth 2. In other words, Knowledge Trees with shallow depth are particularly well suited for complex tasks, in particular for recognizing parts of a sentence that are difficult to identify by logistic regression.

For methods with greater depth, the situation is not so clear: 2 of the 3 results for methods with depth 3 are not statistically significant, and the third result represents the opposite trend: the more accurate the logistic regression, the greater the advantage of Knowledge Trees over logistic regression.

However, the advantage of these methods over logistic regression changes statistically significantly as the complexity of probing parts of the sentence for logistic regression increases.

Therefore, we will determine which probing classifiers perform better when considering only the ``complex'' parts of a sentence, which logistic regression probes with less accuracy.

In this section, we consider a case in which datasets corresponding to parts of a sentence with logistic regression accuracy greater than 98 percent are excluded from the analysis. This is done in order to more closely examine the effectiveness of Knowledge Trees versus logistic regression in probing parts of a sentence in relatively complex cases.

\begin{figure}[ht]
  \centering
  \begin{minipage}{\columnwidth}
    \includegraphics[width=\linewidth]{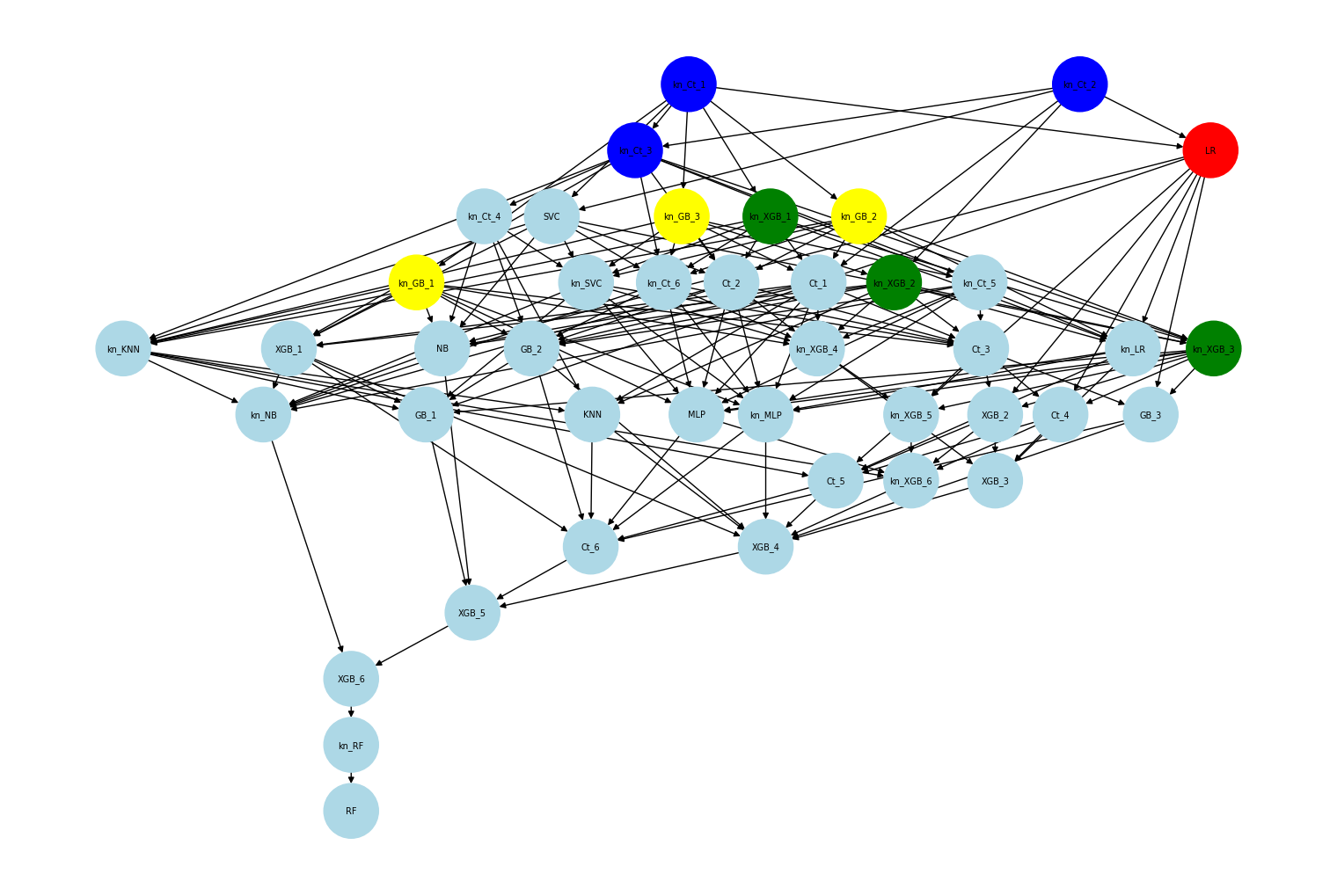}
    \caption{Dominance graph of probing classifiers with following constraint on datasets for different sentence members. Datasets on which logistic regression achieves more than \textbf{98\%} accuracy omitted. Case of relatively difficult classification tasks.}
  \end{minipage}
\end{figure}

Figure 3 shows that when only relatively complex problems for logistic regression are considered, the non-dominated set of probing classifiers consists exclusively of methods from the Knowledge Trees group. Moreover, it includes only CatBoost-based Knowledge Trees of shallow depth, and other methods, including XGBoost-based and Sklearn-based Knowledge Trees, are not included in this set.

\subsection{Dataset size and Knowledge Trees}

Above, we investigated the relationship between the accuracy advantage of different variants of Knowledge Trees over logistic regression and the ``complexity'' of the classification task for logistic regression. We now study the relationship between the advantage of Knowledge Trees over logistic regression in probing parts of sentence and the size of the dataset corresponding to a given part of sentence. Essentially, we want to answer the question on datasets of which size Knowledge Trees perform best and provide the greatest benefits, thereby determining the most appropriate area of its application.

\begin{figure}[ht]
  \centering
  \begin{minipage}{\columnwidth}
    \includegraphics[width=\linewidth]{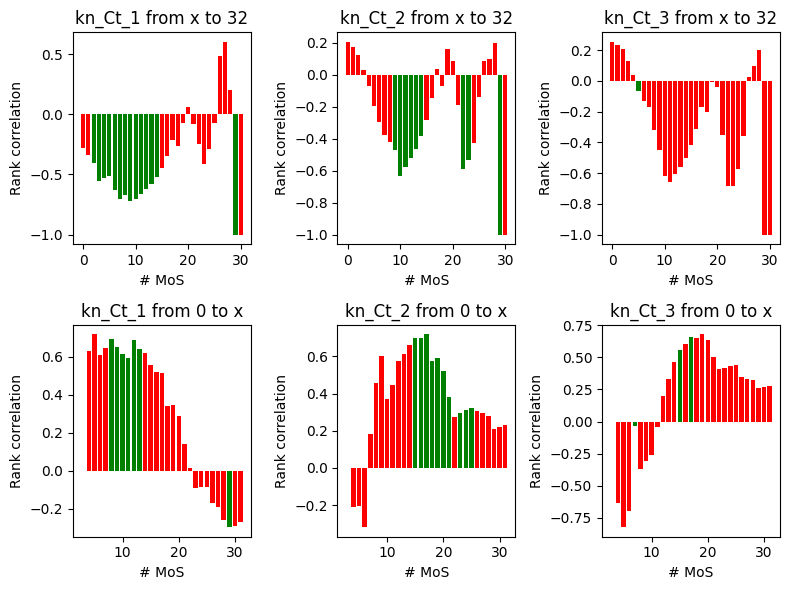}
    \caption{Rank correlation between advantage CatBoost-based Knowledge Trees on logistic regression and size of dataset. At small ranges (pictures in the bottom row) the correlation is positive. At large ranges (pictures in the upper row), the correlation is negative. Results are shown in green statistically significant; in red not statistically significant.}
  \end{minipage}
\end{figure}

Figure 4 clearly illustrates two patterns for CatBoost-based Knowledge Trees of shallow depth (1-2 levels). First, there is a positive rank correlation between dataset size and the advantage provided by Knowledge Trees ranging from smallest to medium datasets. This means that as the size of the dataset increases from small to medium, the advantage of Knowledge Trees over logistic regression increases.

Second, there is a negative rank correlation between the size of the dataset and the advantage of Knowledge Trees in the range of medium to large datasets. This means that as the dataset size increases from medium to large, the advantage of Knowledge Trees over logistic regression decreases.

This result is statistically significant ($p$-value $0.05$) for ensembles of trees of depth 1 and 2, while no clear pattern emerges for depth 3. Calculations show that for XGBoost-based and Sklearn-based Knowledge Trees the patterns are exactly the same and the graphs are remarkably similar, so their graphs are omitted here.

These findings suggest the hypothesis that Knowledge Trees may perform best on medium-sized datasets. We can already argue that Knowledge Trees will outperform logistic regression, but will it outperform other methods?

To verify this, we analyze the accuracy of probing classifiers for different parts of sentence represented by medium-sized datasets. In addition, for contrast and completeness of the study, we will also analyze large datasets, as well as a combined range (medium plus large).

We do not analyze separately the accuracy of probing classifiers on small datasets, since we have already realized that Knowledge Trees perform best for ``complex'' classification tasks, and complex patterns apparently cannot be represented by a small dataset. However, this could be the subject of a separate study.

Thus, in 4.4, we will analyze medium-sized datasets, in subsection 4.5 medium-large datasets, and in 4.6 large datasets.

\subsection{Medium-sized datasets}

In this section, we analyze a case in which probing classifiers are trained on datasets that are under two constraints. First, parts of sentence captured in these datasets are not very accurately (with less than 98\% accuracy) probed by logistic regression. That is, informally speaking, they are ``difficult'' parts of sentence to recognize by logistic regression. Second, only \textit{medium-sized datasets} are considered.

\begin{figure}[ht]
  \centering
  \begin{minipage}{\columnwidth}
    \includegraphics[width=\linewidth]{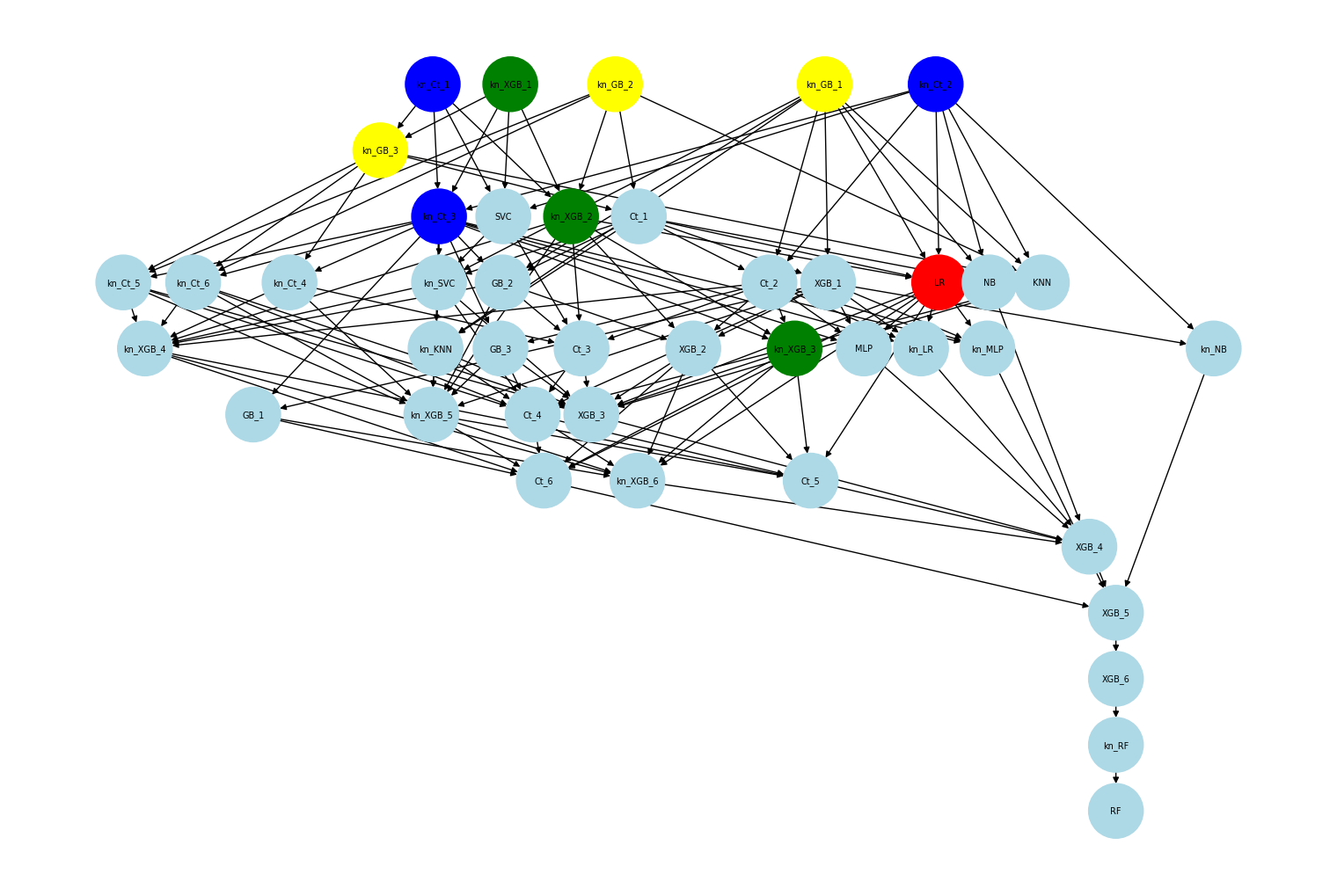}
    \caption{Dominance graph of probing classifiers with following constraint on datasets for different sentence members. Datasets on which logistic regression achieves more than \textbf{98\%} accuracy omitted and only \textit{medium-size} datasets are included.}
  \end{minipage}
\end{figure}

In Fig. 5, we see that the non-dominated set of probing classifiers includes only methods from the Knowledge Trees group, and Logistic Regression (LR) is the dominated method. The composition of the non-dominated set is also interesting: it includes Knowledge Trees of all kinds: two CatBoost-based, two Sklearn-based and only one XGBoost-based. The fact that the improved regularizations in XGBoost do not give a significant advantage over Sklearn-based in this case also requires further explanation.

\subsection{Medium or large datasets}

In this section, we analyze a case in which probing classifiers are trained on datasets that are under two constraints. First, parts of sentence captured in these datasets are not very accurately (with less than 98\% accuracy) probed by logistic regression. That is, informally speaking, they are ``difficult'' parts of sentence to recognize by logistic regression. Second, only \textit{medium or large datasets} are considered.

\begin{figure}[ht]
  \centering
  \begin{minipage}{\columnwidth}
    \includegraphics[width=\linewidth]{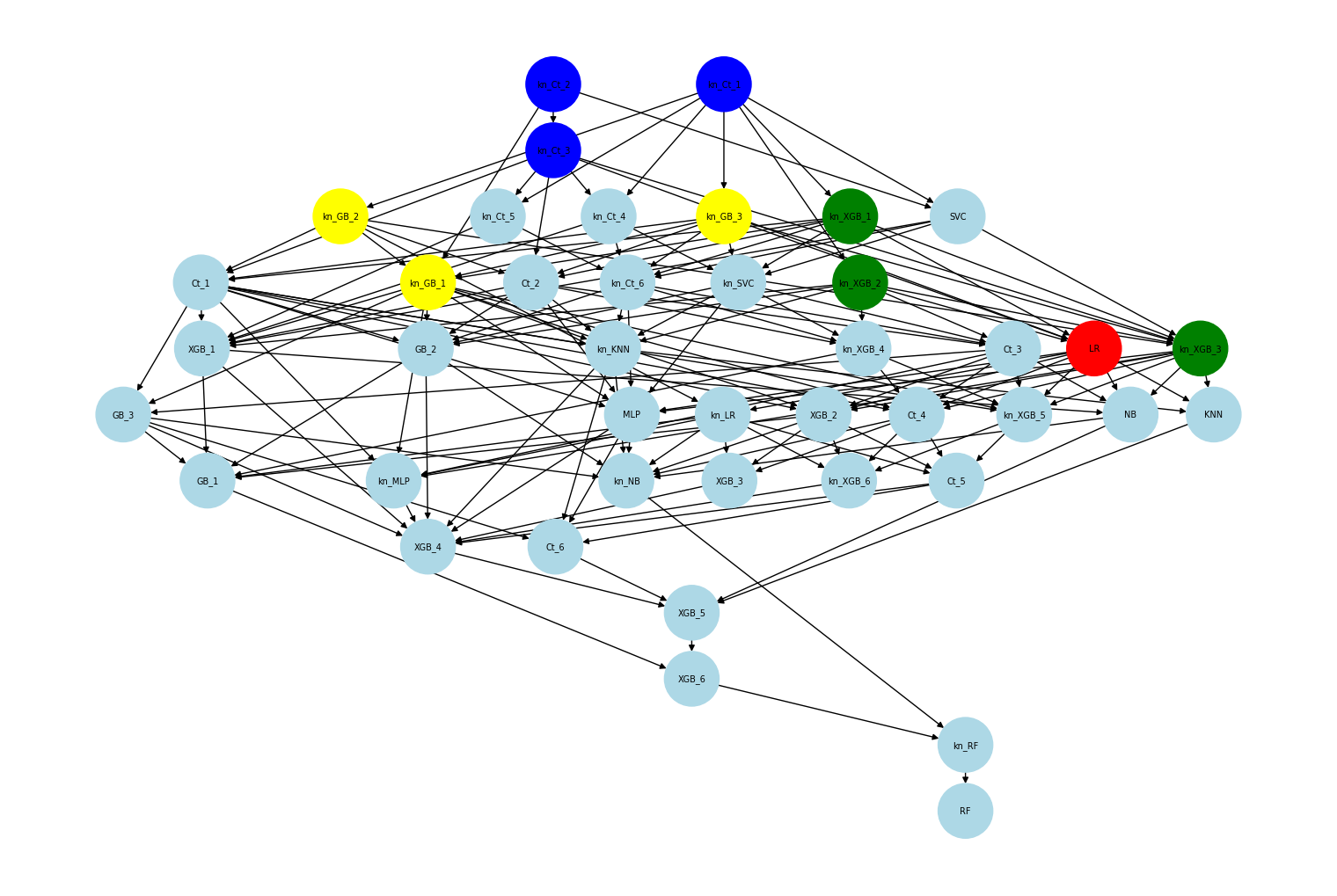}
    \caption{Dominance graph of probing classifiers with following constraint on datasets for different sentence members. Datasets on which logistic regression achieves more than \textbf{98\%} accuracy omitted and only \textit{medium or large size} datasets are included.}
  \end{minipage}
\end{figure}

In Fig. 6, we see that the non-dominated set of methods includes only CatBoost-based Knowledge Trees. This again raises the question of the value of using symmetric trees in gradient boosting.

\subsection{Large datasets}

In this section, we analyze a case in which probing classifiers are trained on datasets that are under two constraints. First, parts of sentence captured in these datasets are not very accurately (with less than 98\% accuracy) probed by logistic regression. That is, informally speaking, they are ``difficult'' parts of sentence to recognize by logistic regression. Second, only \textit{large datasets} are considered.

\begin{figure}[ht]
  \centering
  \begin{minipage}{\columnwidth}
    \includegraphics[width=\linewidth]{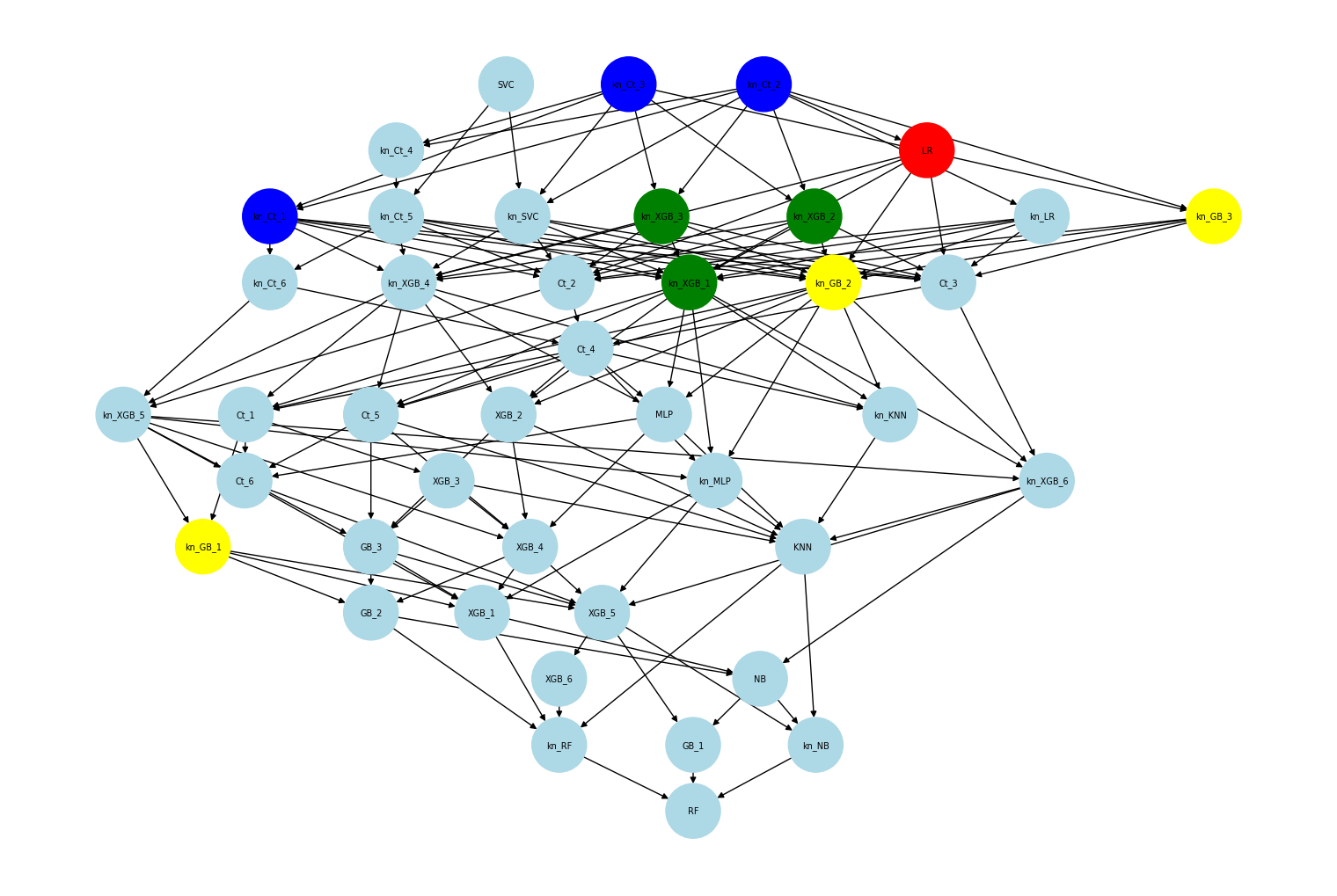}
    \caption{Dominance graph of probing classifiers with following constraint on datasets for different sentence members. Datasets on which logistic regression achieves more than \textbf{98\%} accuracy omitted and only \textit{large-size} datasets are included.}
  \end{minipage}
\end{figure}

Fig. 7 shows that the non-dominated set includes methods from the Knowledge Trees group as well as the Support Vector Machine (SVC) classifier. It is noteworthy that only CatBoost-based ones are present from the Knowledge Trees. It is also interesting to note that in this case study, the depth of the trees in probing classifiers from the non-dominated set is greater than in the previous cases considered, 2 and 3 levels. That is, in this case there are no probing classifiers based on gradient boosting decision stumps in the non-dominated set.

\subsection{General analysis of cases}

We have presented the results of the accuracy analysis of the different probing classifiers for the five cases discussed above in Table 4.

We see that the only method that is constantly in the non-dominated set in all five cases considered is CatBoost-based Knowledge Trees with two levels (kn\_Ct\_2). 

\begin{table}[ht]
\centering
\scriptsize
\begin{tabularx}{\columnwidth}{p{1.5cm}lp{2.1cm}l}
\toprule
Size & Complexity & Prob. classifiers & Advan. \\ 
\midrule
all & 100\% & kn\_Ct\_1, kn\_Ct\_2, kn\_Ct\_3, kn\_XGB\_1, SVC & 18\% \\ 
\hline
all & 98\% & kn\_Ct\_1, kn\_Ct\_2 & 26\% \\ 
\hline
medium & 98\% & kn\_Ct\_1, kn\_Ct\_2, kn\_XGB\_1, kn\_GB\_1, kn\_GB\_2 & 54\% \\ 
\hline
medium+large & 98\% & kn\_Ct1, kn\_Ct2 & 30\% \\ 
\hline
large & 98\% & kn\_Ct2, kn\_Ct3, SVC & 9\%. \\ 
\bottomrule
\end{tabularx}
\caption{Non-dominated set of probing classifiers and the largest statistically significant advantage among Knowledge Trees variants in error rate over logistic regression for cases with different constraints}
\label{your-label2}
\end{table}

In all cases considered, the use of Knowledge Trees is justified. Focusing exclusively on difficult tasks for recognition increases this advantage (from 18\% to 26\%). Additional focus on medium-sized datasets dramatically boosts this advantage from 26\% to 54\%.

Thus, we can conclude that the use of Knowledge Trees is most justified and gives the greatest advantage when training a probing classifier for complex syntactic constructions on a medium-sized dataset.

It can be hypothesized that small datasets are generally unsuitable for learning to probe complex syntactic structures, since they are not enough to convey complex patterns. Thus, the use of Knowledge Trees allows you to switch from using large datasets to medium-sized datasets while maintaining the same level of accuracy, thereby saving time on their compilation.

\section{Discussion}

In all graphs, we observe that methods from the Knowledge Trees group are always part of the non-dominated set of methods. Sometimes they completely fill the non-dominated set, and sometimes they are accompanied by another method, the Support Vector Machine (SVC) based classifier. Logistic regression (LR) and a classifier based on a standard fully connected neural network (MLP) never fall into the non-dominated set. In every scenario we considered, at least one of the Knowledge Trees methods always ended up in the non-dominated set. Thus, it can be argued that Knowledge Trees are an effective tool for probing parts of a sentence.

It is interesting to note that the use of gradient boosting over transformer instead of MLP has already been proposed by \cite{FreeGBDT}. In that work, it was shown that in some cases it is possible to obtain slightly better results than using MLP. This is quite consistent with the results of our study: gradient boosting probing the vector representation of tokens to classify parts of sentence gives results quite comparable to logistic regression and MLP. Somewhere better, somewhere worse.

But if the same gradient boosting probes Knowledge Neurons instead of vector representation of tokens, the results are much better, as our study shows.

In addition, it is interesting to analyze the results of CatBoost-based Knowledge Trees. This, of course, requires additional research, but even from the graphs it is clear that if symmetrical trees were not included in the Knowledge Trees, the results would be much more modest. The frequency of cases where Knowledge Trees will dominate will be less, and even in those cases where they dominate, the degree of this dominance will be lower.

We see that there is not a single case that we have considered where Symmetric Knowledge Trees did not fall into the non-dominated set. On the contrary, in only 2 out of 5 cases Asymmetric Knowledge Trees were included in the non-dominated set. What specific benefits symmetry provides in gradient boosting decision trees requires further study.

It is noteworthy that the Knowledge Stumps have performed well. When a tree has only one level, it cannot be symmetric or asymmetric. This requires further study, but for now we hypothesize that reducing the number of tree levels, even to one, reduces overfitting and hence improves performance, especially for datasets with a small to medium number of examples. Interestingly, for large datasets (as shown in Subsection 4.6), deeper trees, more than one level, become more preferable beyond stumps. This agrees well with the above hypothesis.

In addition, it is interesting to analyze the success of the Support Vector Machine (SVC) classifier. It is the only method, apart from the methods from the Knowledge Trees group, that made it to the non-dominated set of methods in some cases. The success of SVC seems to increase with the number of instances in the dataset, making it competitive. The only significant drawback observed in our study is its extremely slow performance on large datasets. According to our observations, it takes more than ten times longer than methods offering comparable accuracy. However, a comprehensive comparative accuracy analysis is a topic for a separate study.

\section{Conclusion}

This study contributes to ongoing work on syntactic entity recognition in large language models. Using probing by gradient boosting decision trees on hidden layer of feed-forward network of the transformer, we constructed a more efficient probing classifier for syntactic entities compared to logistic regression and other traditional methods. This approach showed a significant reduction in error rate, by 9\% to 54\%, depending on the presets.

Interestingly, the Support Vector Machine Classifier (SVC) also proved to be a notable method in some tasks, suggesting its potential usefulness on datasets with a large number of instances. However, its low learning speed on large datasets is a limitation.

In future research, it is interesting to establish how much less training data Knowledge Trees would require to achieve the same accuracy as traditional methods such as logistic regression. It is also interesting to understand why Symmetric Knowledge Trees are so effective.



\begin{thebibliography}{}

\bibitem{TransformersAndCompositionality}
{\it Dziri, N., Lu, X., Sclar, M., Li, X. L., Jian, L., Lin, B. Y., ... Choi, Y.} (2023). Faith and Fate: Limits of Transformers on Compositionality. arXiv preprint arXiv:2305.18654.

\bibitem{Hallucination}
{\it Rawte, V., Chakraborty, S., Pathak, A., Sarkar, A., Tonmoy, S. M., Chadha, A., ...  Das, A.} (2023). The Troubling Emergence of Hallucination in Large Language Models--An Extensive Definition, Quantification, and Prescriptive Remediations. arXiv preprint arXiv:2310.04988.

\bibitem{LLMandKG}
{\it Pan, S., Luo, L., Wang, Y., Chen, C., Wang, J., Wu, X.} (2023). Unifying Large Language Models and Knowledge Graphs: A Roadmap. arXiv preprint arXiv:2306.08302.

\bibitem{Belinkov}
{\it Belinkov, Y.} (2022). Probing classifiers: Promises, shortcomings, and advances. Computational Linguistics, 48(1), 207-219.

\bibitem{7layer}
{\it Xu, N., Gui, T., Ma, R., Zhang, Q., Ye, J., Zhang, M., Huang, X.} (2022). Cross-Linguistic Syntactic Difference in Multilingual BERT: How Good is It and How Does It Affect Transfer? arXiv preprint arXiv:2212.10879.

\bibitem{Attention}
{\it Vaswani, A., Shazeer, N., Parmar, N., Uszkoreit, J., Jones, L., Gomez, A. N., ... Polosukhin, I.} (2017). Attention is all you need. Advances in neural information processing systems, 30.

\bibitem{UD}
{\it De Marneffe, M. C., Manning, C. D., Nivre, J., Zeman, D.} (2021). Universal dependencies. Computational linguistics, 47(2), 255-308.

\bibitem{KeyValueMemory}
{\it Geva, M., Schuster, R., Berant, J.,  Levy, O.} (2020). Transformer feed-forward layers are key-value memories. arXiv preprint arXiv:2012.14913.

\bibitem{SubUpdates}
{\it Geva, M., Caciularu, A., Wang, K. R.,  Goldberg, Y.} (2022). Transformer feed-forward layers build predictions by promoting concepts in the vocabulary space. arXiv preprint arXiv:2203.14680.

\bibitem{KnowledgeNeurons}
{\it Dai, D., Dong, L., Hao, Y., Sui, Z., Chang, B., and Wei, F.} (2022). Knowledge neurons in pretrained transformers. In Proceedings of the 60th Annual Meeting of the Association for Computational Linguistics (Volume 1: Long Papers), pp. 8493–8502.

\bibitem{XGBoost}
{\it Chen, T., Guestrin, C.} (2016, August). Xgboost: A scalable tree boosting system. In Proceedings of the 22nd acm sigkdd international conference on knowledge discovery and data mining (pp. 785-794).

\bibitem{CatBoost}
{\it Prokhorenkova, L., Gusev, G., Vorobev, A., Dorogush, A. V., Gulin, A.} (2018). CatBoost: unbiased boosting with categorical features. Advances in neural information processing systems, 31.

\bibitem{DistilBERT}
{\it Sanh, V., Debut, L., Chaumond, J., Wolf, T.} (2019). DistilBERT, a distilled version of BERT: smaller, faster, cheaper and lighter. arXiv preprint arXiv:1910.01108.

\bibitem{l1Unpredictability}
{\it Mohammadi, M., Atashin, A. A., Tamburri, D. A.} (2023). From L1 subgradient to projection: A compact neural network for L1-regularized logistic regression. Neurocomputing, 526, 30-38.

\bibitem{FreeGBDT}
{\it Minixhofer, B., Gritta, M., Iacobacci, I.} (2021). Enhancing Transformers with Gradient Boosted Decision Trees for NLI Fine-Tuning. arXiv preprint arXiv:2105.03791.

\end{thebibliography}
\end{document}